\documentclass{article}
\usepackage{ijcai25}

\usepackage{times}
\usepackage{soul}
\usepackage{url}
\usepackage[hidelinks]{hyperref}
\usepackage[utf8]{inputenc}
\usepackage[small]{caption}
\usepackage{graphicx}
\usepackage{amsmath}
\usepackage{amsthm}
\usepackage{booktabs}
\usepackage{algorithm}
\usepackage{algorithmic}
\usepackage[switch]{lineno}
\usepackage{multirow}
\usepackage{subcaption}
\usepackage{amssymb}
\usepackage{color}
\usepackage[table]{xcolor}

\definecolor{orange}{rgb}{1.0,0.5,0.15}

\title{Few-Shot Class-Incremental Model Attribution Using Learnable Representation From CLIP-ViT Features}

\author{
Hanbyul Lee$^1$
\And
Juneho Yi$^2$\thanks{Corresponding author.}\\
\affiliations
$^1$Department of Artificial Intelligence, Sungkyunkwan University\\
$^2$Department of Electrical and Computer Engineering, Sungkyunkwan University\\
\emails
\{byul1748, jhyi\}@skku.edu
}

\begin{document}

\maketitle

\begin{abstract}
    Recently, images that distort or fabricate facts using generative models have become a social concern. To cope with continuous evolution of generative artificial intelligence (AI) models, model attribution (MA) is necessary beyond just detection of synthetic images. However, current deep learning-based MA methods must be trained from scratch with new data to recognize unseen models, which is time-consuming and data-intensive. This work proposes a new strategy to deal with persistently emerging generative models. We adapt few-shot class-incremental learning (FSCIL) mechanisms for MA problem to uncover novel generative AI models. Unlike existing FSCIL approaches that focus on object classification using high-level information, MA requires analyzing low-level details like color and texture in synthetic images. Thus, we utilize a learnable representation from different levels of CLIP-ViT features. To learn an effective representation, we propose Adaptive Integration Module (AIM) to calculate a weighted sum of CLIP-ViT block features for each image, enhancing the ability to identify generative models. Extensive experiments show our method effectively extends from prior generative models to recent ones.
\end{abstract}

\section{Introduction}
\begin{figure*}[t]
    \centering
    \footnotesize
    \begin{tabular}{ccc}
         \includegraphics[width=0.31\linewidth]{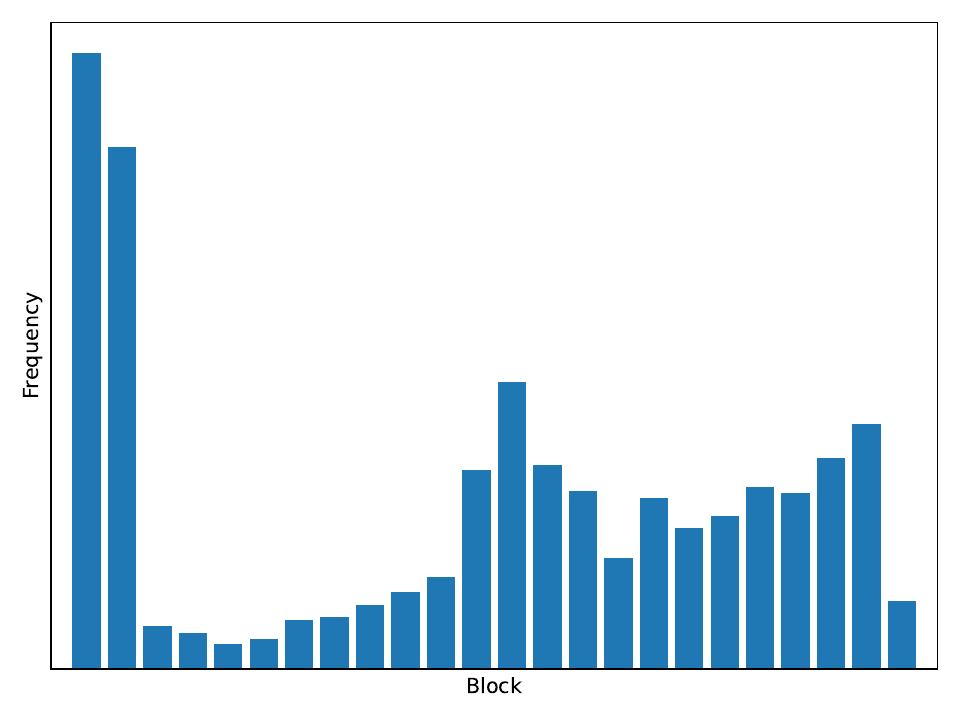} & \includegraphics[width=0.31\linewidth]{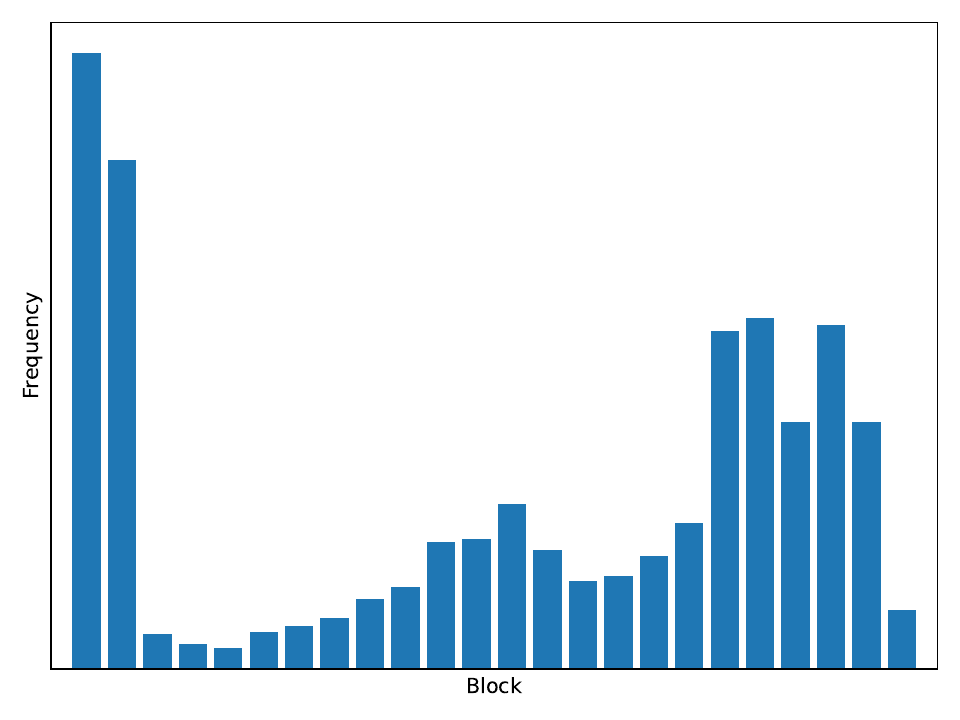}  & \includegraphics[width=0.31\linewidth]{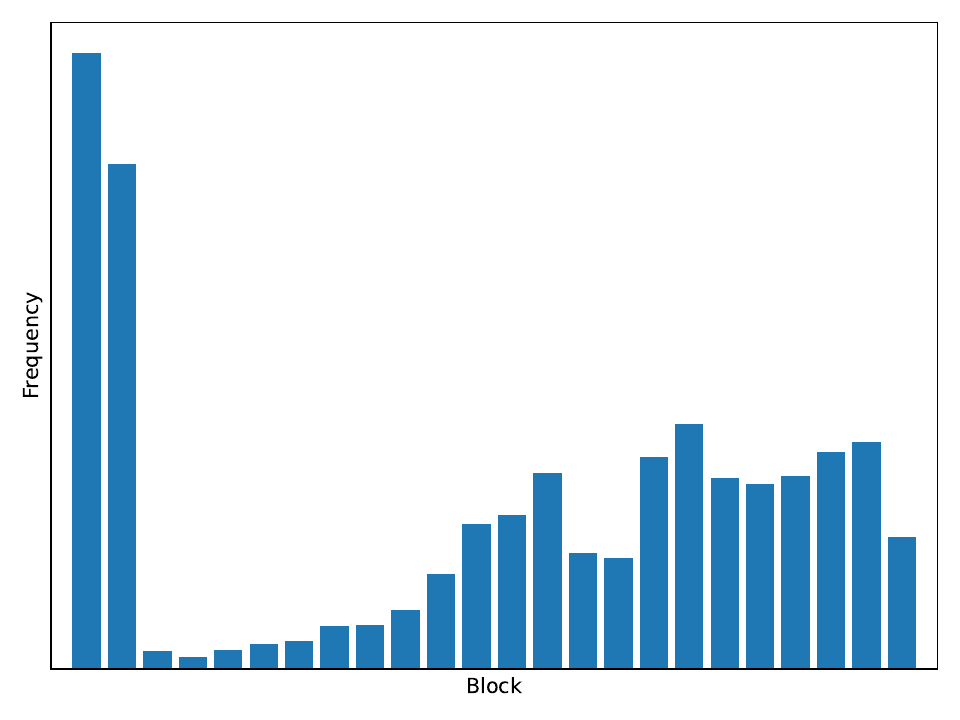}\\
         (a) CycleGAN & (b) DDPM & (c) DALL-E 2 \\
         \includegraphics[width=0.31\linewidth]{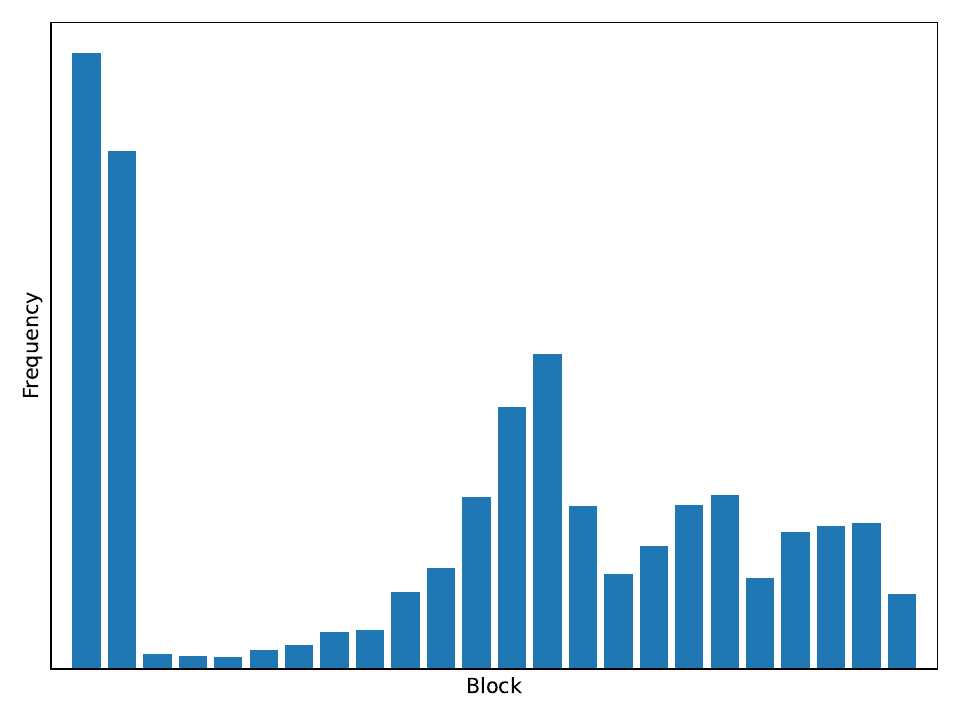} & \includegraphics[width=0.31\linewidth]{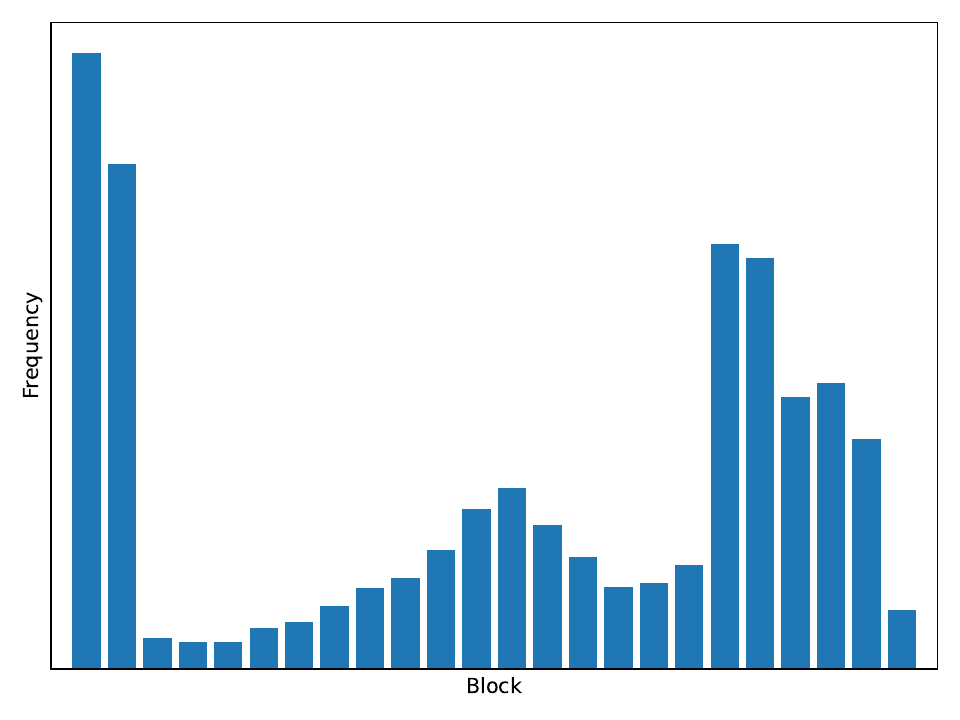} & 
         \includegraphics[width=0.31\linewidth]{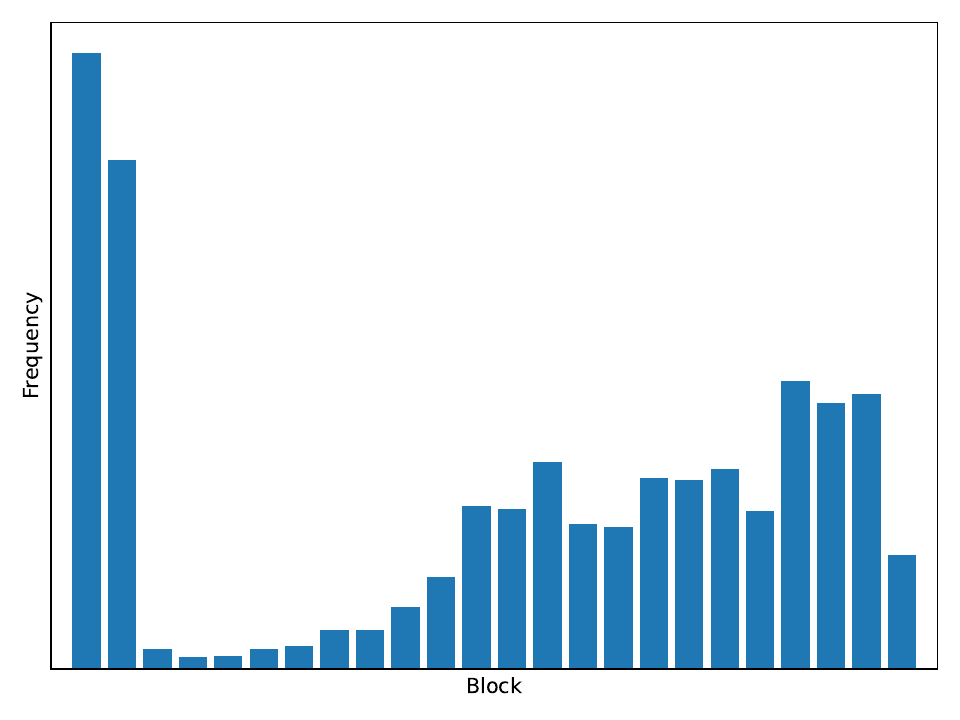} \\
         (d) InfoMaxGAN & (e) Improved Diffusion & (f) Stable Diffusion 1.4\\
    \end{tabular}
    \caption{Frequencies at which each block is ranked as the most important block in MA using the features from Transformer blocks of pre-trained CLIP-ViT. (a) CycleGAN, (b) DDPM, (c) DALL-E 2, (d) InfoMaxGAN, (e) Improved Diffusion, (f) Stable Diffusion 1.4. The frequencies are measured as follows: First, we calculate the weight of each block feature with AIM, then find the block with the largest value and the block with the second largest value for each channel of the weights. If the second largest value is greater than half of the largest value, the frequency of the block with the second largest weight is also counted. Since early blocks of CLIP-ViT extract low-level information and later blocks extract high-level information, these results indicate that the level containing the most important information varies depending on generative models. Specifically, GAN-based models such as CycleGAN and InfoMaxGAN tend to have a high frequency of large weights in the middle blocks. In contrast, models based on more recent structures like DDPM, Improved Diffusion, DALL-E 2, and Stable Diffusion 1.4 have higher frequencies for later blocks than early blocks.}
    \label{fig:frequency}
\end{figure*}
Image generation technology has advanced rapidly with the advent of models such as Variational Autoencoder (VAE)~\cite{kingma2013auto} and Generative Adversarial Network (GAN)~\cite{goodfellow2014generative}. More recently, Diffusion Model (DM)~\cite{ho2020denoising} has made it easier to generate high-quality images, further accelerating the pace of progress.

However, these technological advances have a downside. Malicious users can entirely fabricate images or manipulate existing ones, making it hard to verify their authenticity. This becomes especially problematic when fake images are shared on social networking services (SNS), where they can quickly spread misinformation to a large number of people.

This has led to the need for synthetic image detection (SID). SID is a binary classification task that distinguishes between real images and those generated by various generators. For SID methods, it is important to find characteristics that generated images have in common for generalizability.

In contrast, model attribution (MA) goes beyond simply detecting synthetic images—it identifies the specific model that generated them. In addition to analyzing generated images like SID, MA recognizes characteristics (i.~e., fingerprints) of each generative model. While recent MA approaches have shown their effectiveness~\cite{wang2023where,song2024manifpt}, they remain limited in expandability by their need for extensive retraining on a large amount of data to recognize unseen generators.

To cope with the persistent emergence of unseen generative models, MA methods should be capable of learning from limited data. This capability would enable MA methods to adapt effectively to newly emerging commercial image generation models, whose data is challenging to collect. Furthermore, MA methods should be able to attribute unseen models with minimal additional training rather than full retraining, thereby achieving significant time and cost savings.

Therefore, in this paper, we propose to apply few-shot class-incremental learning (FSCIL) approaches to MA. FSCIL is a learning paradigm in which a trained image classification model is made to classify novel classes by learning a few additional images without forgetting previous classes. To be specific, the training process consists of a base session and incremental sessions. In the base session, the model is trained to classify base classes before seeing novel classes. In the incremental sessions, a few images of unseen classes are given and the model classifies the novel classes by learning them. Consequently, applying the basic concept of FSCIL to MA can effectively cope with unseen generative models.

As to features to use for SID, which is closely related to MA, a prior study~\cite{koutlis2025leveraging} has shown that it is important to consider not only high-level information but also low-level information such as color and texture when analyzing synthetic images. However, since FSCIL has been targeted to object classification, existing FSCIL methods focus on extracting high-level semantic information from images. Thus, direct application of FSCIL to MA does not work well. A representation that includes both low-level and high-level information is needed to achieve an optimal performance in few-shot class-incremental model attribution. To illustrate, Figure~\ref{fig:frequency} shows frequencies at which each block is ranked as the most important block in MA using features from Transformer blocks of pre-trained CLIP-ViT~\cite{radford2021learning}. Since early blocks of CLIP-ViT extract low-level information and later blocks extract high-level information, these results indicate that the feature level containing the most important information varies depending on generative models. As it is not known in advance which level information is effective for MA, we propose Adaptive Integration Module (AIM), which learns the feature weights at various levels. It helps to determine generative models by calculating the optimal weights for each image to perform MA. AIM can be directly applied to FSCIL frameworks which leverage features of pre-trained feature extractors and are not dependent on model structures. Among such FSCIL methods, we use TEEN~\cite{wang2024few}.

A total of 28 image generation models are employed for experiments. We train our proposed framework using only GAN data in the base session, and a small amount of recent generative model data such as DMs in the incremental sessions. We also compare the performance of the proposed method with methods that use only low-level information, methods that use high-level information, and methods that use all levels of information to identify the level that provides important information for MA. Through extensive experiments, we found that methods that utilize low-level information outperform methods that only adopt high-level information generally, and the approaches leveraging all levels of information performing best. Our proposed method showed the best performance across most sessions.

The contributions of this work are as follows:
\begin{itemize}
    \item To cope with the persistent emergence of unseen generative models, we repurpose FSCIL mechanisms for MA problem. 
    \item We propose a learnable representation that can effectively indicate artifacts and fingerprints of a generative model for few-shot class-incremental model attribution. The trainable module, dubbed AIM, enables the optimal use of all levels of features in CLIP-ViT for a given image.
\end{itemize}

\section{Related Works}
\subsection{Synthetic Image Detection}
Synthetic image detection (SID) is the problem of determining whether a given image is generated or real. SID approaches can be divided into low-level information-based methods and high-level information-based methods. As an example of using low-level information, ~\cite{tan2024rethinking} used the fact that adjacent pixels are strongly related to each other because image generation models should perform an up-sample operation while generating an image. On the other hand, ~\cite{ojha2023towards} proposed a high-level information-based approach that leverages the last output feature from pre-trained CLIP-ViT for the detection of synthetic images.

Recently, \cite{koutlis2025leveraging} presented a method called RINE, which utilizes both low-level and high-level information. RINE detects generated images by utilizing all the intermediate output features of pre-trained CLIP image encoder. This method significantly improves SID performance, showing that leveraging all levels of information is effective in analyzing generated images.

\subsection{Model Attribution}
Model attribution (MA) can be seen as an extension of SID to identifying generative models that produced synthetic images. A recent MA study~\cite{song2024manifpt} provided explicit definitions on artifacts and fingerprints of generative models for the first time. Specifically, the difference between real data and generated data in various manifolds such as RGB space, frequency domain, and embedding space of pre-trained models is defined as an artifact, and a fingerprint is defined as a set of such artifacts. In addition, \cite{liu2025model} proposed to perform MA in a few-shot one-class classification setting to address the difficulty of obtaining a sufficient amount of training data for MA in the real world. 

To classify an unseen generative model, most MA methods need to be trained from scratch with images generated by the model whenever it emerges. While the approach that ~\cite{liu2025model} proposed can be extended to multi-class classification through few-shot learning, it is limited in expandability since it is performed in a one-vs-rest method. Therefore, We aim to apply few-shot class-incremental learning to MA to identify new image generation models more efficiently. A recent study~\cite{li2024boosting} explored this few-shot incremental setting for face forgery attribution by employing procedures and training losses that are designed to remove information irrelevant to forgery from high-level semantic features. Unlike their work, we leverage a representation that captures information levels particularly significant for identifying each generative model.

\subsection{Few-Shot Class-Incremental Learning}
Few-shot class-incremental learning (FSCIL) was first proposed by ~\cite{tao2020few}, which allows a trained image classification model to classify new classes by learning only a few additional images. The training process of FSCIL is divided into a base session and incremental sessions. In the base session, the model is trained with a large amount of data like a general deep learning scheme. In the incremental sessions, the model learns a few images of new classes. On the one hand, previous FSCIL studies have used knowledge distillation techniques such as the one proposed by~\cite{park2024pre}. On the other hand,
~\cite{wang2024few} found that when using prototypes obtained from a pre-trained feature extractor, samples of new classes are often misclassified as base classes. Accordingly, they proposed TEEN, which calibrates a prototype of a new class to be close to the prototypes of base classes. 

While it is possible to simply apply these FSCIL methods to MA, the focus of existing FSCIL research has been on high-level information. In tasks like SID and MA, where a fingerprint is less visually apparent than content, it is necessary to consider low-level information as well. Therefore, we adopt a learnable representation that can include both low-level and high-level information.

\begin{figure}[t]
    \centering
    \includegraphics[width=0.8\linewidth]{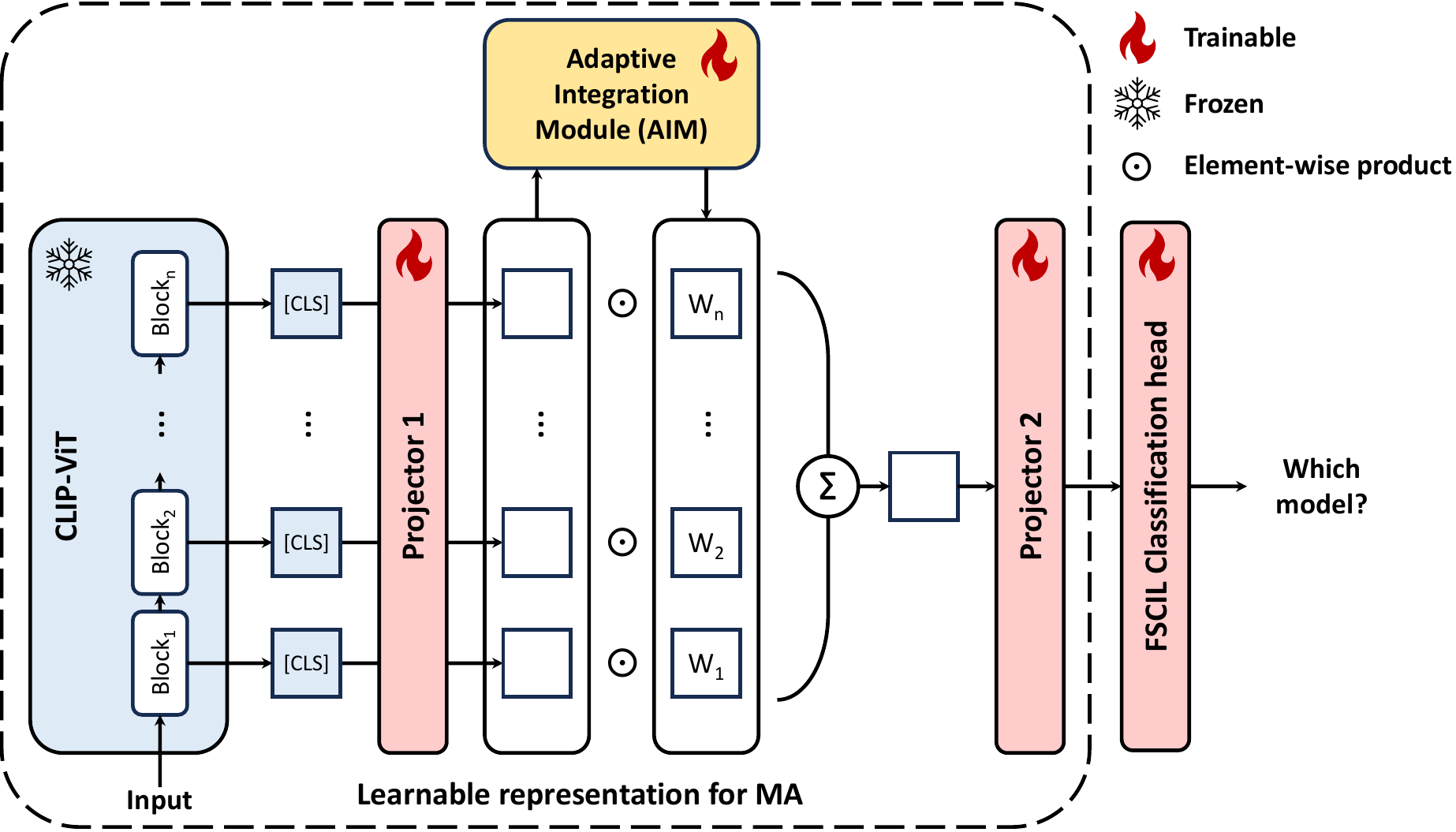}
    \caption{Overview of our proposed method. We utilize a learnable representation for attributing generative models. In the proposed method, all the features from pre-trained CLIP image encoder blocks are taken and integrated. We perform a weighted sum of the features, and the weights used here are obtained using Adaptive Integration Module (AIM). AIM is a trainable module that calculates appropriate weights for a given image.}
    \label{fig:overview}
\end{figure}

\begin{table*}[ht!]
    \centering
    \subfloat[Base session]{\begin{tabular}{c|c}
        \hline
        \textbf{Name} & \textbf{Year} \\
        \hline
        \hline
        BEGAN~\cite{berthelot2017began} & \multirow{5}{*}{2017}\\
        CramerGAN~\cite{bellemare2017cramer} & \\
        ProGAN~\cite{karras2017progressive} &  \\
        CycleGAN~\cite{zhu2017unpaired} & \\
        MMDGAN~\cite{li2017mmd} & \\
        \hline
        SNGAN~\cite{miyato2018spectral} & \multirow{5}{*}{2018}\\
        RelGAN~\cite{nie2018relgan} & \\
        StarGAN~\cite{choi2018stargan} & \\
        BigGAN~\cite{brock2018large} & \\
        GANimation~\cite{pumarola2018ganimation} & \\
        \hline
        S3GAN~\cite{luvcic2019high} & \multirow{5}{*}{2019} \\
        StyleGAN~\cite{karras2019style} & \\
        GauGAN~\cite{park2019semantic} & \\
        STGAN~\cite{liu2019stgan} & \\
        AttGAN~\cite{he2019attgan} & \\
        \hline
        StyleGAN2~\cite{karras2020analyzing} & 2020 \\
        \hline
    \end{tabular}}
    \renewcommand{\arraystretch}{1.31}
    \subfloat[Incremental session]{
    \begin{tabular}{c|c}
        \hline
        \textbf{Name} & \textbf{Year} \\
        \hline \hline
        \cellcolor{lightgray!30} DDPM~\cite{ho2020denoising} & 2020 \\
        \hline
        \cellcolor{lightgray!30} InfoMaxGAN~\cite{lee2021infomax} & \multirow{5}{*}{2021}\\
        \cellcolor{yellow!30}DALL-E~\cite{ramesh2021zero} & \\
        \cellcolor{yellow!30}Improved Diffusion~\cite{nichol2021improved} & \\
        \cellcolor{green!20}Guided Diffusion~\cite{dhariwal2021diffusion} & \\
        \cellcolor{green!20}Glide~\cite{nichol2021glide} & \\
        \hline
        \cellcolor{cyan!10}DALL-E 2~\cite{ramesh2022hierarchical} & \multirow{4}{*}{2022}\\
        \cellcolor{cyan!10}VQ Diffusion~\cite{gu2022vector} & \\
        \cellcolor{violet!20}Stable Diffusion 1.4~\cite{rombach2022high} & \\
        \cellcolor{violet!20}Stable Diffusion 1.5~\cite{rombach2022high} & \\
        \hline
        \cellcolor{pink!30}Midjourney\textsuperscript{1} & \multirow{2}{*}{2023}\\
        \cellcolor{pink!30}Wukong\textsuperscript{2} & \\
        \hline
    \end{tabular}}
    \caption{Detailed composition of generative models, sorted in the chronological order of publication. A total of 28 image generation models are used. We utilize images used in previous studies~\protect\cite{wang2020cnn,ojha2023towards,tan2024rethinking,zhong2023rich}. In the base session (session 1), images from 16 GAN models are employed. Images from the remaining 12 models are adopted in 6 sessions with 2-way 5-shot per incremental session. That is, only GAN images are utilized in the base session and a small amount of recent generative model images are used in the incremental sessions. The models for incremental sessions are employed in \colorbox{lightgray!30}{session 2}, \colorbox{yellow!30}{session 3}, \colorbox{green!30}{session 4}, \colorbox{cyan!10}{session 5}, \colorbox{violet!20}{session 6}, and \colorbox{pink!30}{session 7}.}
    \label{tab:dataset}
\end{table*}

\section{Method}
\subsection{Learnable Representation From CLIP-ViT Features}
\label{sec:representation}
The overall workflow of our proposed method is shown in Figure~\ref{fig:overview}. As mentioned above, since existing FSCIL methods mainly classify images based on high-level (semantic) information, it is necessary to use a representation that includes both low-level and high-level information to achieve optimal performance when applied to MA. Therefore, we utilize output features of every block of pre-trained CLIP image encoder. CLIP-ViT has been trained on large dataset and is known to provide an effective representation when analyzing synthetic images~\cite{ojha2023towards,koutlis2025leveraging}. The output features of every block are integrated through a weighted sum. The weights used in this process are learned through Adaptive Integration Module (AIM). AIM is a trainable module that calculates optimal weights for a given image. The reason for using different weights for each image is because artifacts and fingerprints are stronger at different levels depending on generative models. Also refer to Figure~\ref{fig:frequency}.

The detailed process is as follows. First, a given image $I \in \mathbb{R}^{H \times W \times 3}$ is divided into patches and linearly projected. Then the patch embeddings are fed into frozen CLIP-ViT and $[CLS]$ tokens are taken from each block. The $[CLS]$ token interacts with other patches in ViT~\cite{dosovitskiy2020image} and integrates information from all patches. Thus, a $[CLS]$ token can serve as a representative feature of the output from each block.

The $[CLS]$ token $t^i$ at the output of $i$-th block is a feature vector of dimension $d_0$. Letting the number of blocks in CLIP-ViT be $n$, we can obtain $n$ $[CLS]$ tokens. The first projector $f_1$ projects token $t^i$ as $t_{proj}^{i} \in \mathbb{R}^{d_1}$. AIM then takes the projected tokens as input and outputs a learned per-token weight $W \in \mathbb{R}^{n \times d_1}$. This can be formulated as the following equation:
\begin{equation}
    W = AIM(f_1([t^1, ... , t^n]))
\end{equation}
Then each element of the weight $W$ is scaled between 0 and 1 by applying softmax along the first dimension:
\begin{equation}
    W_{scaled}^{i} = \frac{e^{W^i}}{\sum_{j=1}^{n}e^{W^j}}
\end{equation}
$W^i \in \mathbb{R}^{d_1}$ indicates the weight for the $i$th $[CLS]$ token. The weighted sum $t_{integrated} \in \mathbb{R}^{d_1}$ of the tokens $[t_{proj}^1, ..., t_{proj}^n]$ is calculated as follows:
\begin{equation}
    t_{integrated} = \sum_{i=1}^{n} W_{scaled}^{i} \odot t_{proj}^i
\end{equation}
$\odot$ denotes element-wise multiplication. We feed this integrated feature $t_{intergrated}$ into the second projector $f_2$ and get the final output feature.

\begin{table*}[t]
    \centering
    \begin{tabular}{c|c||c|c|c|c|c|c|c}
        \hline
         \multirow{2}{*}{\textbf{Info level}} & \multirow{2}{*}{\textbf{Method}} & \multicolumn{7}{c}{\textbf{Session}} \\
         \cline{3-9}
         & & \textbf{Base (1)} & \textbf{2} & \textbf{3} & \textbf{4} & \textbf{5} & \textbf{6} & \textbf{7} \\
         \hline \hline
         
         \multirow{3}{*}{\textbf{Low}} & NPR~\cite{tan2024rethinking} & 93.44 & 91.67 & 87.35 & 81.75 & 77.35 & 71.39 & 67.38 \\
         & ManiFPT$_{RGB}$~\cite{song2024manifpt} & 84.34 & 81.58 & 74.80 & 69.16 & 64.44 & 58.73 & 55.86 \\
         & ManiFPT$_{Freq}$~\cite{song2024manifpt} & 89.47 & 85.94 & 80.00 & 73.25 & 67.42 & 63.19 & 59.52 \\
         \hline
         \multirow{3}{*}{\textbf{High}} & UnivFD~\cite{ojha2023towards} & 72.84 & 57.14 & 57.03 & 52.86 & 52.85 & 48.81 & 46.21 \\
         & ManiFPT$_{SL}$~\cite{song2024manifpt} & 51.88 & 43.56 & 40.18 & 37.41 & 33.94 & 30.98 & 29.11 \\
         & ManiFPT$_{SSL}$~\cite{song2024manifpt} & 75.34 & 67.47 & 62.48 & 57.11 & 52.17 & 47.75 & 44.07 \\
         \hline
         \multirow{2}{*}{\textbf{Low \& High}} & RINE~\cite{koutlis2025leveraging} & 97.47 & \textbf{96.92} & 91.15 & 84.16 & 78.98 & 73.12 & 69.29 \\
         & Ours & \textbf{98.06} & 96.72 & \textbf{92.70} & \textbf{85.82} & \textbf{81.31} & \textbf{74.75} & \textbf{70.59} \\
         \hline
    \end{tabular}
    \caption{Accuracy (\%) when using each method. The best performance is highlighted in bold. We can see that methods based on low-level information show higher performance than the methods based on high-level information. These results confirm that low-level information does indeed have a significant impact when analyzing generated images. It is also observed that the performance is maximized when both levels of information are utilized. In particular, our proposed method shows the highest performance in most sessions.}
    \label{tab:comparison_result}
\end{table*}

\subsection{Exploiting FSCIL Framework for Model Attribution}
The feature extracted as described in Section~\ref{sec:representation} can be directly applied to FSCIL frameworks that are not dependent on model structures. In particular, we employ TEEN~\cite{wang2024few}.

TEEN is a prototype-based FSCIL method that calibrates new class prototypes in incremental sessions to be closer to the prototypes of base classes, based on the fact that new classes are often misclassified as the base classes. This is a training-free method where no adjustments are made in the base session, and they are made only in the incremental session. Let the prototype of a new class be $t_{n}$, and the prototype $\bar{t_{n}}$ after calibration is given by

\footnotetext[1]{\url{https://www.midjourney.com/home/}}
\footnotetext[2]{\url{https://xihe.mindspore.cn/modelzoo/wukong}}

\begin{equation}
    \bar{t_{n}} = \alpha t_{n} + (1-\alpha)\Delta t_{n}
\end{equation}
where $\alpha$ is a hyperparameter which controls how much the prototype calibration affects $t_{n}$. Also $\Delta t_{n}$ consists of a weighted sum of base prototypes and indicates how much $t_n$ should be calibrated. Given the number of base classes, $B$, and the base prototype, $t_b$, we can denote $\Delta t_{n}$ as
\begin{equation}
    \Delta t_{n} = \sum_{b=0}^{B-1}w_{b,n}t_{b}
\end{equation}
The weight $w_{b,n}$ for each base prototype is determined by the distance between $t_b$ and $t_n$. The distance is measured by cosine similarity $S_{b,n}$ of the two prototypes $t_b$ and $t_n$:
\begin{equation}
    S_{b,n} = \frac{t_{b} \cdot t_{n}}{||t_{b}|| \cdot ||t_{n}||} \cdot \tau
\end{equation}
$\tau$ is a hyperparameter that adjusts the sharpness of the weight distribution. Therefore, the weights $w_{b,n}$ can be written as
\begin{equation}
    w_{b,n} = \frac{e^{S_{b,n}}}{\sum_{j=0}^{B-1}e^{S_{j,n}}}
\end{equation}
To sum up, the calibrated prototype $\bar{t_{n}}$ is given by
\begin{equation}
    \bar{t_{n}} = \alpha t_{n} + (1-\alpha)\sum_{b=0}^{B-1}w_{b,n}t_{b}
\end{equation}

In this way, we perform few-shot class-incremental model attribution based on TEEN as a FSCIL framework using an effective representation for MA.

\section{Experiments}
\label{sec:experiments}
\subsection{Setting and Implementation}
We use images generated from 28 generative models used in the previous studies~\cite{wang2020cnn,ojha2023towards,tan2024rethinking,zhong2023rich}. The details on the models are summarized in Table~\ref{tab:dataset}. In the base session, we adopt data from 16 models for training. Images from the remaining 12 models are used in 6 incremental sessions with 2-way 5-shot. In other words, the MA model is trained using only GAN images in the base session, and a small number of images from recent generative models such as autoregressive models and DMs are considered for the incremental sessions. This setting enables us to confirm whether an MA method is expandable from prior models to recent models.

As aforementioned, we employ TEEN~\cite{wang2024few} with ViT-L/14 as the backbone of CLIP image encoder. However, our method is applicable to any FSCIL method that leverages pre-trained backbone models. The code for TEEN is adopted from the official repository\textsuperscript{3}. The hyperparameters for TEEN training are set to  $\alpha=0.5$ and $\tau=16$ with a learning rate 0.001 and batch size 128. All learnable modules are trained using cross-entropy loss. The two projectors and AIM consist of two blocks including Linear, Dropout, and ReLU layers. All experiments are conducted using Python 3.10, Pytorch 2.2 on a single NVIDIA A100 GPU.

\footnotetext[3]{\url{https://github.com/wangkiw/TEEN}}

\begin{figure*}[ht!]
    \centering
    \begin{tabular}{ccccc}
          \includegraphics[width=0.12\linewidth]{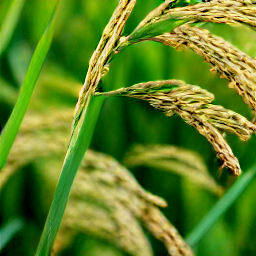} & \hspace{-0.3cm}\includegraphics[width=0.16\linewidth]{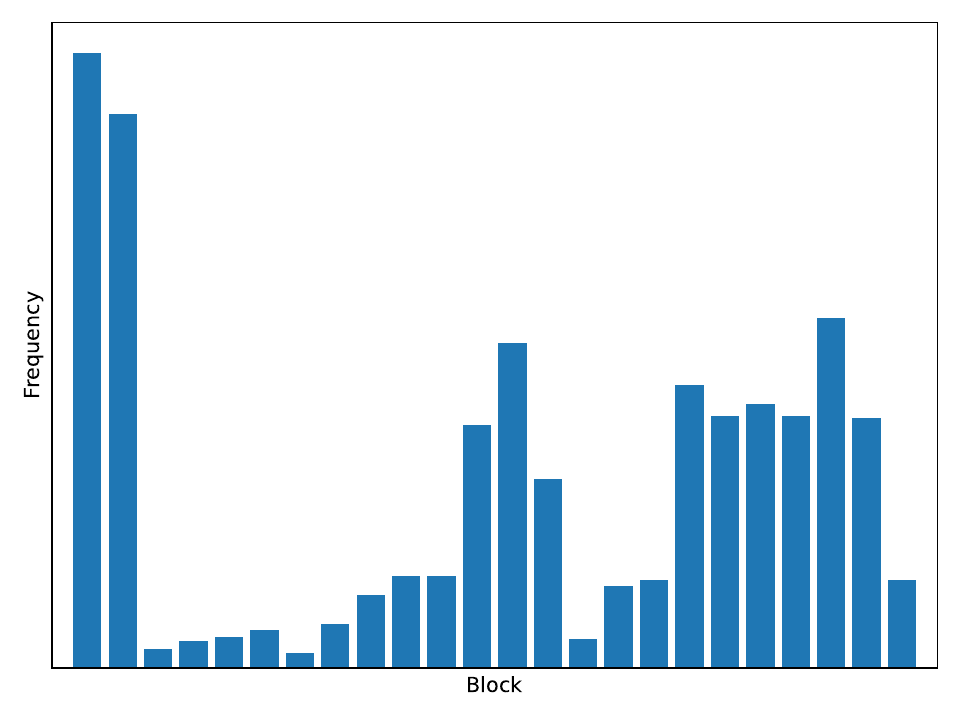} & \includegraphics[width=0.12\linewidth]{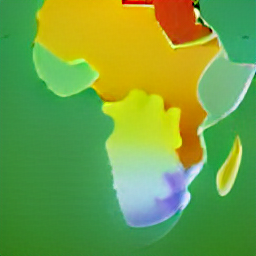} & \hspace{-0.3cm}\includegraphics[width=0.16\linewidth]{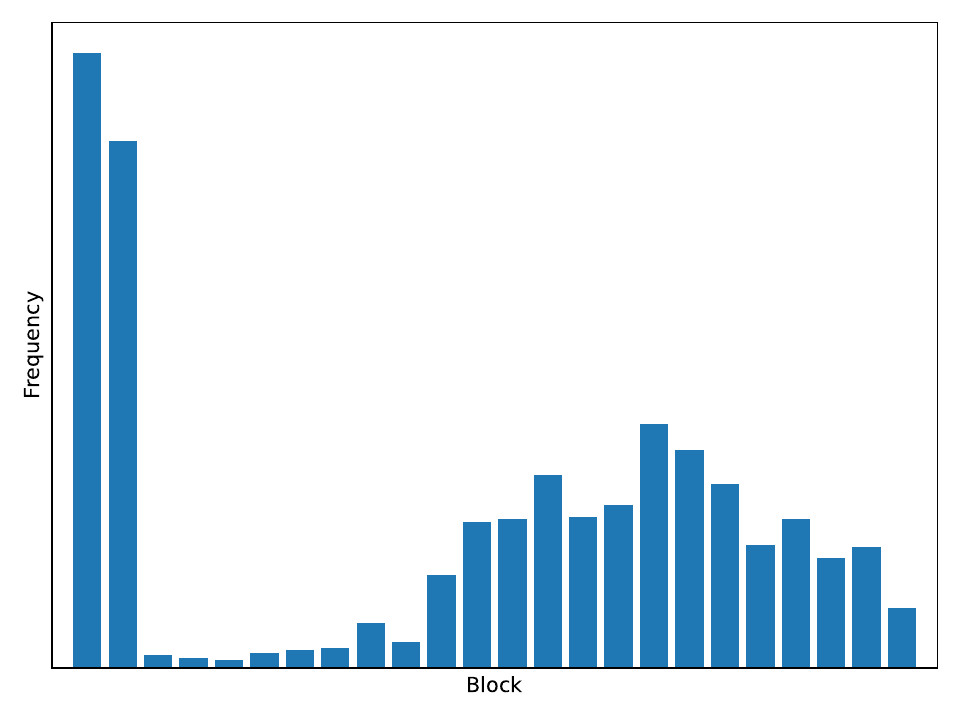} & \multirow{4}{*}[0.65in]{\includegraphics[width=0.3\linewidth]{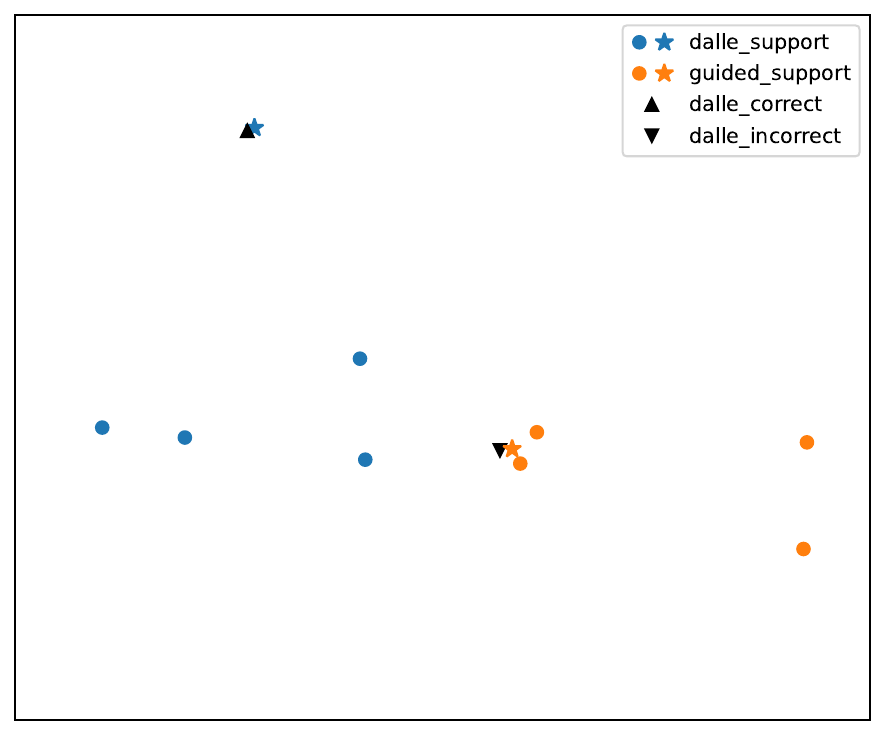}}  \\
          \multicolumn{2}{c}{$\blacktriangle$ DALL-E test image (correct)} & \multicolumn{2}{c}{\textcolor{blue}{\large{\textbf{$\star$}}} DALL-E support image} & \\
          \vspace{-0.2cm}\\
          \includegraphics[width=0.12\linewidth]{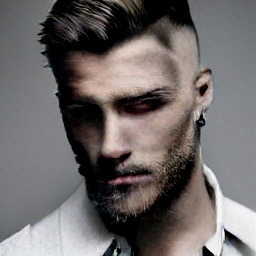} & \hspace{-0.3cm}\includegraphics[width=0.16\linewidth]{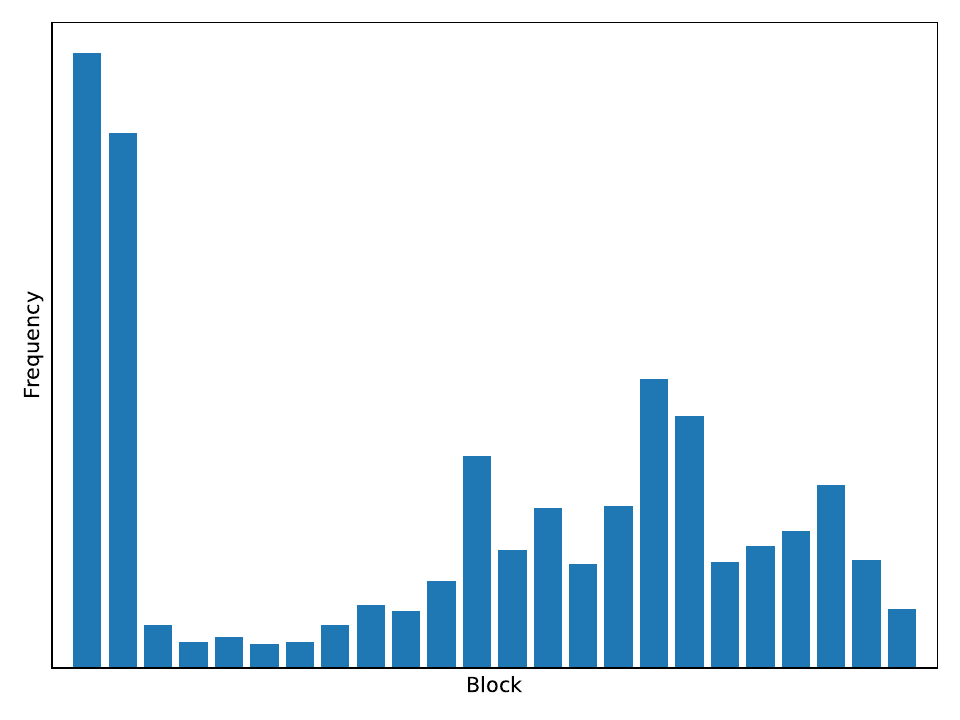} & \includegraphics[width=0.12\linewidth]{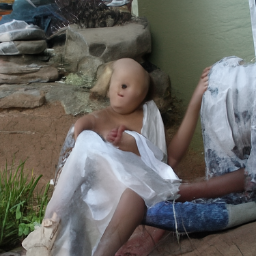} & \hspace{-0.3cm}\includegraphics[width=0.16\linewidth]{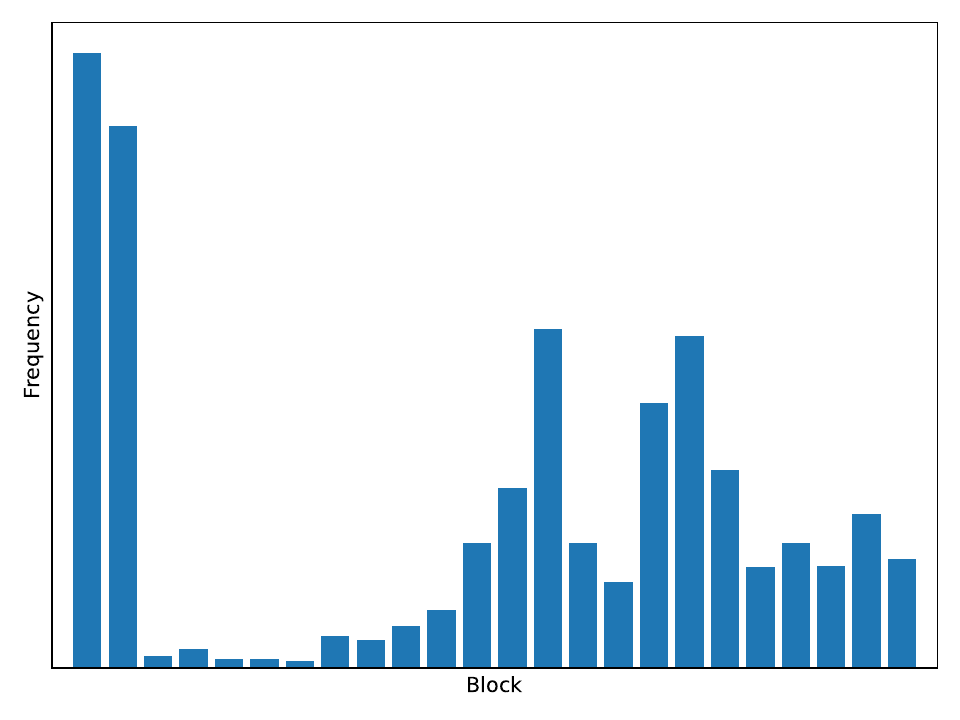} & \\
          \multicolumn{2}{c}{$\blacktriangledown$ DALL-E test image (incorrect)} & \multicolumn{2}{c}{\textcolor{orange}{\large{\textbf{$\star$}}} Guided Diffusion support image} & (b) t-SNE result of CLIP low-level features\\
          \vspace{-0.3cm}\\
          \multicolumn{4}{c}{(a) Correct and incorrect attribution cases} & \\
    \end{tabular}
    \caption{Correct and incorrect cases in attributing images generated by DALL-E. (a) shows sample images and frequencies of CLIP-ViT blocks that are considered the most important for each image. The frequencies are obtained in the same way as Figure~\ref{fig:frequency}. $\blacktriangle$ means a correctly attributed DALL-E test image and $\blacktriangledown$ is incorrectly attributed to Guided Diffusion. \textcolor{blue}{$\star$} and \textcolor{orange}{$\star$} indicate an image in the DALL-E and Guided Diffusion support sets, respectively. (b) is the result of t-SNE performed on CLIP-ViT low-level features of DALL-E support set, Guided Diffusion support set, and DALL-E test images. As aforementioned and shown in (a), attribution results are highly affected by features from early blocks. Consequently, if a low-level representation of an image generated by a particular model is similar to those of other generative models, it may lead to incorrect attribution.}
    \label{fig:good_bad}
\end{figure*}

\subsection{Information Level for Model Attribution}
\label{sec:layer_importance}
\subsubsection{Analysis on Effective Information Level}
\label{sec:information_level}
We analyze the effectiveness of using both low-level and high-level information, and which level of information is particularly important for different models. For this purpose, we count the frequencies that a block has the largest value in the weights learned by AIM. The results are summarized and shown in Figure~\ref{fig:frequency}. We calculate the weights for each block feature with AIM. Then we find the block with the largest value and the block with the second largest value for each channel of the weights. Along with the frequency of the block with the largest value, the frequency of the block with the second largest value is also counted if its weight is greater than half of the largest value. In this experiment, we use ViT-L/14 as the backbone of CLIP image encoder. Note that ViT-L/14 consists of 24 Transformer blocks in total.

As can be seen from the results, low-level information is necessary for identifying generative models and other important level of information varies by model structure. It is observed that the information from the first and second blocks is the most important across the models. As other important level of information, GAN-based models such as CycleGAN and InfoMaxGAN show a high frequency of large weights in middle blocks. On the other hand, models based on more recent structures such as DDPM, Improved Diffusion, DALL-E 2, and Stable Diffusion 1.4 show higher frequencies in later blocks than in earlier blocks, meaning that high-level information tends to be more important. 

Therefore, in order to effectively apply existing FSCIL methods to MA, which mainly utilize high-level information, it is necessary to use a representation that includes multiple levels of information together, and our proposed AIM that can learn the optimal weights for each given image.

\subsubsection{Performance by Information Level}
To validate the observations above, we compare the performance of different methods in cases of using low-level information, high-level information, and both levels of information. We adopt the following four methods to compare performance:
\begin{itemize}
    \item \textbf{Low-level}:  NPR~\cite{tan2024rethinking}, ManiFPT$_{RGB}$, ManiFPT$_{Freq}$~\cite{song2024manifpt}
    \item \textbf{High-level}:  UnivFD~\cite{ojha2023towards}, ManiFPT$_{SL}$, ManiFPT$_{SSL}$~\cite{song2024manifpt}
    \item  \textbf{Low \& High}: RINE~\cite{koutlis2025leveraging}
\end{itemize}
NPR is a method based on low-level information that detects synthetic images using a resulting image of four adjacent pixels minus the value of one pixel as input. UnivFD is a SID method that exploits high-level information using only the last output feature of CLIP-ViT. ManiFPT is an MA method that defines the difference between generated data and real data on a specific manifold as an artifact and uses it for attributing models. ManiFPT in RGB space or frequency domain is classified as a low-level information method, and ManiFPT which uses an embedding space of supervised learning (SL) or self-supervised learning (SSL) models is categorized as a high-level information method. For experiments, we adopt pre-trained ResNet50~\cite{he2016deep} as a supervised learning model and Barlow Twins~\cite{zbontar2021barlow} as a self-supervised learning model for high-level ManiFPT following the previous study~\cite{song2024manifpt}. RINE, like the proposed method, utilizes all block outputs from CLIP-ViT but integrates the features with a single learned weight for all input images.

Both UnivFD and RINE are based on CLIP using ViT-L/14 and the same pre-trained parameters as our proposed method. The classifier for NPR and ManiFPT is ResNet50. In terms of implementation, we refer to official repositories of NPR\textsuperscript{4}, UnivFD\textsuperscript{5}, and RINE\textsuperscript{6}. For ManiFPT, we develop our own implementation based on the description in the paper~\cite{song2024manifpt}.

\footnotetext[4]{\url{https://github.com/chuangchuangtan/NPR-DeepfakeDetection}}
\footnotetext[5]{\url{https://github.com/WisconsinAIVision/UniversalFakeDetect}}
\footnotetext[6]{\url{https://github.com/mever-team/rine}}

The results are presented in Table~\ref{tab:comparison_result}. Overall, the low-level information-based methods are more accurate than the high-level methods. These results show that low-level information does play an important role when detecting synthetic images. It is also observed that the performance is maximized when both levels of information are utilized. In particular, our proposed method shows the highest performance in most sessions. It is noteworthy that the proposed method maintains high performance even though it only learns GAN images in the base session and learns a small amount of data from the recent generative models afterward. This indicates that the proposed method is indeed expandable from previous to latest unseen models.

As discussed earlier, MA results tend to be strongly influenced by low-level features extracted from early blocks of CLIP-ViT. To illustrate, Figure~\ref{fig:good_bad} shows correct and incorrect cases when our method is applied to DALL-E images. The sample images and frequencies at which each block feature is identified as the most important are presented in Figure~\ref{fig:good_bad} (a). They demonstrate that features from the first and second blocks of CLIP-ViT are critical across all samples. In Figure~\ref{fig:good_bad} (b), we visualize t-SNE~\cite{van2008visualizing} result of low-level features of the support and test images. The test images are attributed to generative models whose support images have similar low-level features to those of the test images. Specifically, when a DALL-E test image is correctly attributed to DALL-E, its low-level representation is close to those of the DALL-E support images. Conversely, if its low-level representation is closer to the Guided Diffusion support set than the DALL-E support set, a DALL-E test image is incorrectly attributed to Guided Diffusion. 

\begin{table*}[ht!]
    \centering
    \begin{tabular}{c|c||c|c|c|c|c|c|c}
        \hline
         \multirow{2}{*}{\textbf{Backbone}} & \multirow{2}{*}{\textbf{Method}} & \multicolumn{7}{c}{\textbf{Session}} \\
         \cline{3-9}
         & & \textbf{Base (1)} & \textbf{2} & \textbf{3} & \textbf{4} & \textbf{5} & \textbf{6} & \textbf{7} \\
         \hline \hline
         \multirow{3}{*}{\textbf{ViT-B/32}} & UnivFD~\cite{ojha2023towards} & 71.50 & 56.31 & 56.38 & 51.68 & 50.85 & 48.04 & 45.36 \\
         & RINE~\cite{koutlis2025leveraging} & 86.31 & 85.22 & 79.13 & 72.07 & 67.10 & 62.71 & 59.23 \\
         & Ours & \textbf{87.06} & \textbf{85.72} & \textbf{80.48} & \textbf{74.52} & \textbf{69.83} & \textbf{64.85} & \textbf{61.18} \\
         \hline
         \multirow{3}{*}{\textbf{ViT-L/14}} &UnivFD~\cite{ojha2023towards} & 72.84 & 57.14 & 57.03 & 52.86 & 52.85 & 48.81 & 46.21 \\
         & RINE~\cite{koutlis2025leveraging} & 97.47 & \textbf{96.92} & 91.15 & 84.16 & 78.98 & 73.12 & 69.29 \\
         & Ours & \textbf{98.06} & 96.72 & \textbf{92.70} & \textbf{85.82} & \textbf{81.31} & \textbf{74.75} & \textbf{70.59} \\
         \hline
    \end{tabular}
    \caption{Accuracy (\%) when changing the backbone of CLIP-based model attribution methods (ViT-B/32 and ViT-L/14). The best performance for each backbone is highlighted in bold. The proposed method correctly attributes models regardless of the backbone. RINE and the proposed method, which use features from all blocks, show higher performance than UnivFD, which employs only the last output feature. Furthermore, our proposed AIM improves performance by using appropriate weights for each image.}
    \label{tab:ablation}
\end{table*}

\begin{table*}[ht]
    \centering
    \begin{tabular}{c|c||c|c|c|c|c|c|c}
        \hline
         \multirow{2}{*}{\textbf{\# of base classes}} & \multirow{2}{*}{\textbf{Prior/Recent}} & \multicolumn{7}{c}{\textbf{Session}} \\
         \cline{3-9}
         & & \textbf{Base (1)} & \textbf{2} & \textbf{3} & \textbf{4} & \textbf{5} & \textbf{6} & \textbf{7} \\
         \hline \hline
         \multirow{2}{*}{\textbf{8}} & Prior & 97.50 & 94.85 & 87.04 & 78.04 & 70.97 & 62.53 & 59.15 \\
         & Recent & 99.69 & 96.25 & 91.04 & 80.36 & 72.50 & 64.31 & 58.98 \\
         \hline
         \multirow{2}{*}{\textbf{12}} & Prior & 98.29 & 96.61 & 90.00 & 81.69 & 76.05 & 69.09 & 64.50 \\
         & Recent & 98.54 & 97.64 & 93.94 & 84.53 & 78.55 & 71.68 & 66.81 \\
         \hline
         \textbf{16} & Full & 98.06 & 96.72 & 92.70 & 85.82 & 81.31 & 74.75 & 70.59 \\
         \hline
    \end{tabular}
    \caption{Accuracy (\%) when varying the composition of base classes. The lower number of base classes leads to higher performance in the early sessions, but we observe a significant decrease in performance towards the later sessions. Furthermore, we see that using recent models when composing base classes has a positive impact.}
    \label{tab:ablation_num_base}
\end{table*}

\subsection{Comparison on CLIP-based Representation}
\label{sec:clip_representation}
In this experiment, we conduct comparison on different backbone configurations of CLIP ViT-B/32 and ViT-L/14. Specifically, UnivFD~\cite{ojha2023towards} uses only the output feature of the last block, while RINE~\cite{koutlis2025leveraging} integrates the representations of all blocks based on a single learned weight for all input images. Our proposed method exploits all block features as RINE, but during the integration process, we use AIM to calculate the optimal weights for a given image.

The results are shown in Table~\ref{tab:ablation}. The proposed method classifies the models accurately regardless of the backbone. Compared to UnivFD, which uses only the last output feature, RINE and the proposed method which leverages features from all blocks show higher performances. Furthermore, our proposed AIM further improves the performance by using learned weights for each image.

\subsection{About Composition of Base Classes}
We analyze the effect of the number of base classes and their composition on performance in the incremental sessions. We first compare the performance when the number of base classes is 8, 12, or 16 (full). We also observe the performance utilizing the prior model images and the recent ones for training when using only a subset of the full base classes. The prior models are selected in the order in Table~\ref{tab:dataset} (a) from the top, and the recent models are selected from the bottom. The detailed compositions of the base classes for each subset are as follows: 
\begin{itemize}
    \item 8 classes (prior): \{BEGAN, CramerGAN, ProGAN, CycleGAN, MMDGAN, SNGAN, RelGAN, StarGAN\}
    \item 8 classes (recent): \{BigGAN, GANimation, S3GAN, StyleGAN, GauGAN, STGAN, AttGAN, SytleGAN2\}
    \item 12 classes (prior): \{BEGAN, CramerGAN, ProGAN, CycleGAN, MMDGAN, SNGAN, RelGAN, StarGAN, BigGAN, GANimation, S3GAN, StyleGAN\}
    \item 12 classes (recent): \{MMDGAN, SNGAN, RelGAN, StarGAN, BigGAN, GANimation, S3GAN, StyleGAN, GauGAN, STGAN, AttGAN, SytleGAN2\}
\end{itemize}
Note that the composition of novel classes in the incremental sessions is not changed from the setting of Table~\ref{tab:dataset} (b) across the experiments.

The results are summarized in Table~\ref{tab:ablation_num_base}. The lower the number of base classes, the higher the performance in the early sessions. However, the performance decreases significantly in the later sessions. Furthermore, it is observed that using recent models as base classes has a positive effect on performance. This is because the more recent the model, the more similar its components (e.~g., training dataset, learning method, etc.) are to those of the latest models. Therefore, the proposed method is better able to recognize characteristics that are important for classifying new classes based on what it has learned from the base classes.

\section{Conclusions}
In this paper, we have repurposed FSCIL frameworks for MA problem to cope with the persistent emergence of unseen generative models. We have also proposed an effective learnable representation for few-shot class-incremental model attribution. Based on the fact that early blocks of CLIP-ViT encode low-level information and later blocks encode high-level information, we have integrated the features from all blocks from pre-trained CLIP-ViT. When integrating the features, Adaptive Integration Module (AIM) has learned the optimal weights for a given image and we perform a weighted sum of the features using the weights. In the experiments, we have shown that the attribution performance can be improved by using an appropriate representation rather than directly applying the existing FSCIL method to MA. We have also confirmed that the proposed method can be continuously expandable by learning a small number of images of the recent generative models, even if it is trained only on previous GAN images.

\appendix

\section*{Acknowledgments}
This work was supported in part by Institute of Information \& communications Technology Planning \& Evaluation (IITP) grant funded by the Korea government (MSIT) (RS-2019-II190421, SKKU Artificial Intelligence Graduate School Program) and the High-performance Computing (HPC) Support project funded by the Ministry of Science and ICT and National IT Industry Promotion Agency (NIPA).

\bibliographystyle{named}
\bibliography{ijcai25}

\end{document}